\documentclass[12pt, a4paper]{article}

\usepackage[T1]{fontenc}
\usepackage[utf8]{inputenc}
\usepackage{float}
\usepackage{authblk}
\usepackage{amsmath, amsthm, amssymb}

\usepackage[margin=1in]{geometry}
\usepackage{microtype}

\usepackage{booktabs}
\usepackage{graphicx}
\usepackage{array}
\usepackage{hyperref}

\usepackage{natbib}
\usepackage{hyperref}
\hypersetup{colorlinks=true, linkcolor=blue, citecolor=blue, urlcolor=blue}

\newtheorem{observation}{Observation}[section]
\theoremstyle{definition}

\theoremstyle{remark}

\title{Fast Cubical Persistent Homology on 2D and 3D Images via Union-Find, Pruning, and Lookup Tables}
\author[1,2]{Titouan Le Breton\thanks{\texttt{titouan.lebreton17@gmail.com}}}
\author[1]{Karol Szustakowski}
\author[1,3]{Marie Piraud}

\affil[1]{Helmholtz AI, Helmholtz Munich, Neuherberg, Germany}
\affil[2]{École des Ponts ParisTech, ENS Paris-Saclay}
\affil[3]{Institute of AI for Health, Helmholtz Munich, Neuherberg, Germany}
\date{}

\begin{document}

\maketitle

\begin{abstract}
We present Flash Cubical, a highly efficient computation of cubical persistence on a V-filtration for 2D and 3D images over~$\mathbb{F}_2$.
The implementation is built around three core ideas.
First, cubical complexes satisfy properties that allow for the computation of persistence of the highest dimension via union-find and duality. 
Second, pruning of certain edges allows for a fast and efficient implementation of union-find. 
Third, the use of a lookup table, which exploits the regularity of cubical complexes to pre-compute local information. This avoids the need to compute local information at run time.
To the best of our knowledge, this is the most efficient implementation of cubical persistence with a V-filtration, both in terms of time and memory costs.
Although the paper focuses on persistence for V-filtration cubical complexes, the underlying ideas generalise naturally to T-filtrations on cubical complexes and suggest promising directions for other complexes.
\end{abstract}

\section{Introduction}
\label{sec:intro}

Persistent homology has become a standard tool for topological data analysis, providing a multiscale summary of the topology of a dataset in the form of a persistence diagram or barcode \citep{edelsbrunner2002, zomorodian2005}.
Cubical complexes, which are the natural representation of grayscale images and volumetric data, are one of the prominent domains for persistence computation.
Cubical filtrations can be built from images in two ways:  V-filtrations and T-filtrations.
The V-filtration assigns the value of each pixel to a vertex and the value of cells is determined by the maximum value of their facets, whereas the T-filtration assigns the value of each pixel to a top-dimensional cell (square in 2D, cube in 3D) and derives the value of a cell as the minimum of its cofacets. 

Several software packages address cubical persistent homology. 
\citet{wagner2012} demonstrated efficient computation via standard  matrix reduction for cubical data.
CubicalRipser \citep{kaji2020} provides a high-performance
implementation for cubical persistence.
The GUDHI library \citep{maria2014} includes a cubical persistence module.
The homology/cohomology duality framework \citep{desilva2011} underlies several reduction strategies across these tools.
Despite this progress, there is still room for improvement in the computation of cubical persistence. This work addresses this gap with three complementary ideas 
and their integration in a benchmarked C++ implementation.
The ideas themselves are not restricted to this setting: they generalise to T-filtrations, with promising directions for simplicial filtrations, notably alpha filtrations.

\paragraph{Contributions.}
\begin{enumerate}
  \item An observation that, if a complex has top-dimensional facets with 1 or 2 co-facets, then the persistence can be computed via union-find on a dual graph. A supporting argument is given in
  Appendix~\ref{app:proof}.
  Notably, this allows for computation of the top dimension of any cubical filtration derived from an image.
  \item A local pruning strategy that eliminates zero-persistence and already-classified cells before running union-find, reducing both sorting cost and the number of "find" operations.
  \item A lookup table that pre-computes local information,
  reducing the work that has to be done at runtime.
  \item A benchmarked C++ implementation combining these ideas, demonstrated on synthetic and real 2D and 3D images. The implementation is available at \url{https://github.com/T-prog123/FlashCubical/tree/main}.
\end{enumerate}

\section{Preliminaries}
\label{sec:prelim}

This section fixes notations and terminology for persistent homology, cubical filtrations,
and the ordering conventions used in the algorithms below.

\subsection{Persistent Homology}

Given a filtration $\emptyset = K_0 \subseteq K_1 \subseteq \cdots\subseteq K_m = K$, persistent homology tracks the birth and death of topological features. We work with coefficients in $\mathbb{F}_2$. 
The standard matrix reduction algorithm \citep{edelsbrunner2002,
zomorodian2005} processes cells in filtration order and identifies \emph{persistence pairs} $(\sigma, \tau)$, where $\sigma$ is \emph{positive} (it creates a new homology class) and $\tau$ is \emph{negative} (it destroys one).
The group $H_p$ denotes homology in dimension $p$; equivalently, $p$-dimensional features are born from positive $p$-cells and killed by negative $(p+1)$-cells. Thus $H_0$ pairs vertices with edges and tracks connected components, $H_1$ pairs edges with squares and tracks loops or tunnels, and $H_2$ pairs squares with cubes and tracks voids.

\subsection{Cubical Complexes}

Cubical complexes are cell complexes adapted to data defined on a regular grid. They are built from elementary cells: vertices, edges, squares, and, in 3D, cubes. More generally, each cell can be viewed as a product of elementary intervals, some degenerate and some non-degenerate, and its dimension is the number of non-degenerate factors. A cubical complex is a collection of such cells that is closed under taking boundaries: whenever a cell belongs to the complex, all of its boundary cells also belong to the complex.

Under the V-filtration, pixels map to vertices. A cell 
$\sigma$ enters the filtration when all of its vertices have appeared, so that
\[
  f_V(\sigma) = \max_{v \in \sigma} f_V(v).
\]
This is the usual lower-star filtration on cubical complexes \citep{robins2011}..

Under the T-filtration, pixel values map to top-dimensional cells (squares in 2D, cubes in 3D). A lower-dimensional cell $\sigma$ enters as soon as one of its incident top-dimensional cells has appeared, so that
\[
  f_T(\sigma)
  =
  \min_{\substack{\tau \supseteq \sigma \\ \dim \tau = d}} f_T(\tau),
\]
where $d$ is the dimension of the complex.

In both cases, a persistence pair $(\sigma,\tau)$ has zero-persistence when the two paired cells have the same filtration value, i.e. $f(\sigma) = f(\tau).$

\subsection{Filtration Order and Tie-Breaking}
\label{sec:filtration_order}

To compute persistence, a \textit{fully refined} filtration is required: at each step of the filtration, exactly one cell appears. This induces a strict total order on the simplexes. 
The V-filtration induces only a partial order on simplices, since two cells can enter the filtration at the same time: as such, it is not fully refined. There are two situations which can lead to two cells appearing at the same time:
\begin{itemize}
    \item Two pixels have the same values. These leads to two vertices entering the filtration at the same time. We resolve this with a \textit{first-order tie-break}.
\item Multiple higher dimensional cells being introduced alongside a vertex. When a vertex $v$ is introduced in the filtration, its entire lower star $L(v)$ [\cite{robins2011}] is introduced with it. This is resolved via a \textit{second-order tie-break}.
\end{itemize}

To compare two given cells of same dimension $\sigma_1$ and $\sigma_2$, we first need to find the vertices $v_1$ and $v_2$ such that $\sigma_1\in L(v_1)$ and $\sigma_2\in L(v_2)$,  and then compare in lexicographic order:
$$(v_1 \text{ vs } v_2, \text{ first order tie-break},  \text{ second order tie-break})$$
Each tie-break is used only at the level where the previous comparison is tied; otherwise it is neither needed nor defined. 
The first and second-order tie-break rule are used to define the fully refined filtration on which persistence is computed.

The \textit{semi-refined} filtration is defined as the filtration that uses the first order tie-break but not the second order tie break.

Importantly, when discussing the zero-persistence cases when implementing pruning and lookup tables (Section \ref{sec:tricks_pruning} and Section \ref{sec:tricks_lookup}) we are referring to zero-persistence in the semi-refined filtration.

Zero-persistence pairs that derive from the first-order tie-break are more complicated to handle than zero-persistence pairs that derive from the second-order tie-break; we therefore do not rely on the local simplification to remove them. 

The algorithm implemented (see Section~\ref{sec:algorithm}) still targets the original V-filtration: any pair whose birth and death value is the same in the original filtration is omitted from the reported diagram.

\section{Theoretical and Computational Principles}
\label{sec:tricks}

This section presents the three ideas that allow for a fast and efficient computation of cubical persistence on a V-filtration. Although we focus on this specific setting, the ideas extend beyond, as we note in each subsection. 

\subsection{Dual Union-Find for Top-Dimensional Persistence}
\label{sec:tricks_dualuf}

We begin by observing that if each top dimensional facet (edges in 2D, squares in 3D) has one or two co-facets (squares in 2D, cubes in 3D), the persistence of the top dimension can be computed via union-find on a dual graph (we define it formally later in this subsection) instead of an expensive matrix reduction. Notice that this does not just apply to cubical complexes, but potentially to other filtrated complexes such as alpha simplicial complexes. 
For the rest of this section, we focus on cubical complexes and on an implementation that works both for V and T-filtrations. 

This dual union-find trick is the most effective in lower dimensional complexes, where computing the top dimension represents a very significant part of the overall computational cost. In 2D, for example, all persistence computation can be done via union-find. In 3D, both $H_0$ and $H_2$ can be computed via union-find, leaving just a simplified $H_1$ to be computed, where we already know which edges and squares are irrelevant because they are paired in another dimension.

If the cubical complex is of dimension $d$, the top dimensional co-facets (of dimension $d$) are mapped to a dual vertex, and the top dimensional facets (of dimension $d-1$) are mapped to dual edges connected to their co-facets/dual vertex. We add an additional "infinite" vertex, and any facets with one co-facet are also connected to this "infinite" vertex. This is a cohomology setting for persistence computation, so dual edges/facets are added in the reverse order with respect to the original fully refined filtration. The unique "infinite" vertex is the first to appear in this dual filtration order.

By running union-find on this dual graph, we get persistence pairs in the reverse order. Since we are in the cohomology setting, dual positive cells found are negative cells (in the traditional homology sense), and the dual negative cells are positive cells (in the traditional homology sense). 

\vspace{0.2cm}

In the cubical setting, the computation of the top dimension via union-find on a dual graph can be seen as a consequence of the theory developed in \citep{bleile2022, garin2020} on dual cubical complexes. Whereas the paper used this theory to prove a link between T and V-filtration, it can also be used to illustrate that the top dimension of a cubical complex maps to the lower dimension of its dual complex, meaning this top dimension can be computed via union-find.

However, as mentioned previously, we believe that this union-find duality idea extends beyond cubical complexes. As such, in Appendix~\ref{app:proof}, we give elements of a proof on why we think this idea goes beyond just cubical complexes and might be used in other settings to also improve performance. 

\subsection{Edge Pruning via Local Pair Detection}
\label{sec:tricks_pruning}

Before union-find is computed, two classes of edges can be
identified and eliminated, substantially reducing sorting and
processing cost and making the union-find algorithm faster:
\begin{itemize}
    \item Zero-persistence direct negative edges:  In the semi-refined filtration, an edge $e$ is a \emph{direct
    zero-persistence negative} edge if it is paired with one of its two vertices and has zero-persistence in the semi-refined filtration.
    \item Positive edges: a positive edge is not relevant when computing the persistence. During the union-find, it just leads to two costly "find" operations that don't have further purpose. As such, they can be removed.
\end{itemize}

The terminology used here applies to the  union-find over $H_0$ (as used in \cite{bauer2021} and \cite{kaji2020}). But this pruning optimization can be applied in exactly the same way when computing the top dimension via dual union-find. We just need to set the context of a dual graph, with dual edges and dual vertices, as well as being in a cohomology context, with a reversed filtration order and a swapped positive/negative meaning. 

In practice, we prune zero-persistence positive edges (in the semi-refined filtration) since these are easier to detect, meaning non-zero-persistence positive edges still have to be processed by union-find. There is an exception for this in 2D, where an additional optimization we call \textit{cross-pruning} is used. In 2D, the union-find algorithm for $H_0$ and $H_1$ use the same edges. As such, once we have computed $H_0$, we know which are the dual positive edges (negative in the classical homology sense), so they can all be pruned before computing $H_1$ (including dual positive edges that are not zero-persistence). We compute $H_0$ first because it has the most persistence pairs, so there will be a minimal number of positive edges where the "find" operation is run for nothing. 

This pruning strategy applies to union-find in general, so by extension it works for both T and V cubical complex filtrations. However, for non-cubical complexes (for example $H_0$ in a VR or alpha simplicial complex), finding these edges to prune might be more costly (in the cubical case, finding these edges is cheap because a lookup table is used, see Section~\ref{sec:tricks_lookup}). A more efficient  alternative is to focus on \textit{apparent pairs}, which are a specific subgroup of these pairs [\cite{bauer2021}, \cite{zhang2020}], which are much easier to detect and are a type of zero-persistence direct pairs. 

\subsection{Local Precomputed Lookup Tables}
\label{sec:tricks_lookup}
The fundamental motivation behind the use of a lookup table is that cubical complexes are regular, with a pre-determined configuration, and such regularity should be leveraged to its fullest extent. Cubical complexes are much more regular than general simplicial complexes derived from point clouds, and we believe that regularity should translate into a specialized but more efficient algorithm. Needless to say, this method does not extend beyond cubical complexes, or at least requires strong regularity. 

The use of a lookup table in a V-filtration to exploit such regularity is based in two key observations:
\begin{itemize}
    \item For a given vertex, its entire lower star $L(v)$ (as defined in Section~\ref{sec:filtration_order}) and any information relative to the second-order tie-break is entirely determined by which of its neighbouring vertices were present in the semi-refined filtration before it. 
    \item There is a finite number of configurations for already-present neighbouring vertices.
\end{itemize}

By neighbouring vertex, we mean a vertex that shares a top-dimensional cell, meaning there are 8 neighbouring vertices in 2D, 26 in 3D.


For each cell, the following local information is precomputed: 
\begin{itemize}
    \item classification: whether the cell is locally positive or negative (locally meaning paired with another cell in $L(v)$) , zero-persistence (in the semi-refined filtration), direct (as defined in Section~\ref{sec:tricks_pruning})
    \item the second-order tie-break ordering of this cell, i.e its order among the cells of the same dimension also introduced with the vertex (so also belonging to $L(v)$)
\end{itemize}

Precomputed here means that these only need to be computed once, stored to disk, and never computed again. For a cubical complex filtration, for any given vertex, the neighbourhood of vertexes present before that vertex and, as such, $L(v)$ and everything that derives from it must fall into one of these precomputed configurations.

The number of configurations that have to be precomputed is much smaller than what it might seem at first glance. A baseline estimate is that there are $2^{26}\approx67$ million possible configurations in 3D. However, for a given vertex $v$, note that if the vertex above $v$ is absent, then the upper left or upper right vertex being absent or not doesn't change anything in this computation, i.e. these configurations collapse to being the same. This generalized to the idea that, if directly adjacent vertex are absent, diagonal vertex being present or not does not matter, so the configurations collapse to a single state. With this consideration, we are faced with just 15 935 possible configurations in 3D.

For a T-filtration, this notion can be adapted by using the lower star of a top dimensional cell, which can be defined as all the lower-dimensional cells that appear along with the top-dimensional cell when it enters the filtration. 


\section{Algorithm and Implementation}
\label{sec:algorithm}

We now describe how the computational ideas from
Section~\ref{sec:tricks} are assembled into complete algorithms to compute persistence on 2D and 3D V-filtrations. In both cases, the computation starts by associating to each vertex a local configuration: the set of neighbouring vertices that appear
before it in the semi-refined filtration. This configuration is used to query the precomputed lookup table.

\subsection{The 2D Algorithm}
\label{sec:algorithm_2d}

In two dimensions, persistence in both dimensions can be computed using union-find. The algorithm combines a union-find computation for \(H_0\) with a dual union-find computation for \(H_1\).

The computation proceeds as follows.

\begin{enumerate}
    \item For each vertex, compute the 8-bit mask of neighbouring vertex preceeding it in the filtration and query the 2D lookup table. This identifies the second order tie-break ordering, direct zero-persistence pairs, and the cells that remain to be processed.

    \item Use the lookup table to identify (zero-persistence) positive edges in \(H_1\). These edges cannot be negative for \(H_0\), so they are removed from the \(H_0\) union-find computation.
    Also, direct zero-persistence edges that are negative are directly applied for $H_0$, those that are positive (so dual negative) are directly applied for $H_1$

    \item Compute \(H_0\) by union-find on the remaining edges, processed  in filtration order
    
    \item Remove the \(H_0\)-negative edges from the \(H_1\) dual union-find computation (cross-pruning)

    \item Compute \(H_1\) by dual union-find 
\end{enumerate}

Hence the 2D pipeline is
\[
\text{lookup}
\;\longrightarrow\;
\text{local pruning}
\;\longrightarrow\;
H_0\text{ union-find}
\;\longrightarrow\;
\text{cross-pruning}
\;\longrightarrow\;
H_1\text{ dual union-find}.
\]

\subsection{The 3D Algorithm}
\label{sec:algorithm_3d}

In three dimensions, the bottom and top homology dimensions are computed by union-find. The middle dimension, \(H_1\), is then computed by implicit reduction after removing cells paired in the $H_0$ or $H_2$.

The computation proceeds as follows.

\begin{enumerate}
    \item For each vertex, compute the 26-bit mask of neighbouring vertex preceeding it in the filtration and query the 3D lookup table. 

    \item Compute \(H_0\), by pruning with the lookup table then union-find,

    \item Compute \(H_2\) by pruning with the lookup table then dual union-find.

    \item Prepare the reduced \(H_1\) problem from the cells not already paired in $H_0$ or $H_2$. Edges paired in \(H_0\) are removed from the row set, squares paired in \(H_2\) are removed from the column set. We also avoiding computing matrix reduction on any zero-persistence \(H_1\) pairs detected by the lookup table. 

    \item Compute the remaining \(H_1\) pairs by matrix reduction over \(\mathbb{F}_2\).
\end{enumerate}

Thus the 3D pipeline is
\[
\resizebox{\textwidth}{!}{$
\text{lookup}
\;\longrightarrow\;
H_0\text{ prune + union-find}
\;\longrightarrow\;
H_2\text{ prune + dual union-find}
\;\longrightarrow\;
\text{pruned }H_1\text{ reduction}
$}
\]

\section{Experiments and Results}
\label{sec:experiments}

\subsection{Main V-filtration benchmarks}

All benchmarks were run on with an Intel Core
i7-12650H processor and 16GB of RAM, using single-threaded execution.
Lookup tables are built once, beforehand.  Synthetic inputs are i.i.d. uniform variables on $[0,1]$.  Real 2D inputs are the Lena image and the DIV2K dataset averaged over the first five samples.  Real 3D inputs are the Fuel, Bonsai, and Aneurism volumetric datasets,
available from \url{http://klacansky.com/open-scivis-datasets/category-all.html}. FlashCubical is compared with Cubical Ripser ([\cite{kaji2020}] version 0.0.33) and Gudhi ([\cite{maria2014}]version 3.12.0).

Table~\ref{tab:main_v_benchmarks} reports output-equivalent
V-filtration benchmarks.  Across all tested cases, FlashCubical is the most efficient, both in terms of memory and time. FlashCubical has the most efficiency gains relative to the benchmarks on 3D data and on synthetic data. 

\begin{table}[H]
\centering
\small
\setlength{\tabcolsep}{6pt}
\caption{Main V-filtration benchmark results. Time is reported in
seconds; peak memory is reported in MB.}
\label{tab:main_v_benchmarks}
\begin{tabular}{llrr}
\toprule
Case & Package & Time (s) & Peak MB \\
\midrule
\multicolumn{4}{l}{\textit{Synthetic 2D medium ($316 \times 316$)}} \\
& Flash   & \textbf{0.02} & \textbf{3}    \\
& GUDHI   & 0.12 & 26   \\
& CRipser & 0.04 & 16   \\
\midrule
\multicolumn{4}{l}{\textit{Lena ($256 \times 256$)}} \\
& Flash   & \textbf{0.01} & \textbf{1}    \\
& GUDHI   & 0.08 & 17   \\
& CRipser & 0.02 & 14   \\
\midrule
\multicolumn{4}{l}{\textit{Synthetic 2D large ($1024 \times 1024$)}} \\
& Flash   & \textbf{0.11} & \textbf{23}   \\
& GUDHI   & 1.95 & 266  \\
& CRipser & 0.34 & 208  \\
\midrule
\multicolumn{4}{l}{\textit{DIV2K ($1024 \times 1024$, avg.\ $n=5$)}} \\
& Flash   & \textbf{0.09} & \textbf{22}   \\
& GUDHI   & 1.34 & 209  \\
& CRipser & 0.27 & 208  \\
\midrule
\multicolumn{4}{l}{\textit{Synthetic 3D medium ($46^3$)}} \\
& Flash   & \textbf{0.06} & \textbf{11}   \\
& GUDHI   & 0.39 & 44   \\
& CRipser & 0.31 & 37   \\
\midrule
\multicolumn{4}{l}{\textit{Fuel ($64^3$)}} \\
& Flash   & \textbf{0.03} & \textbf{24}   \\
& GUDHI   & 0.59 & 193  \\
& CRipser & 0.28 & 53   \\
\midrule
\multicolumn{4}{l}{\textit{Synthetic 3D large ($128^3$)}} \\
& Flash   & \textbf{2.21}  & \textbf{239}  \\
& GUDHI   & 13.79 & 958  \\
& CRipser & 8.29  & 783  \\
\midrule
\multicolumn{4}{l}{\textit{Bonsai ($128^3$)}} \\
& Flash   & \textbf{0.53} & \textbf{194}   \\
& GUDHI   & 7.28 & 1384  \\
& CRipser & 2.94 & 440   \\
\midrule
\multicolumn{4}{l}{\textit{Aneurism ($128^3$)}} \\
& Flash   & \textbf{0.28} & \textbf{189}   \\
& GUDHI   & 6.16 & 1519  \\
& CRipser & 2.62 & 437   \\
\bottomrule
\end{tabular}
\end{table}

\subsection{Reference comparisons}

The following comparisons are between algorithms that do not produce the same output, so they are treated separately.  

We evaluate how much runtime and memory usage can be reduced by computing only two of the three 3D persistences, thereby sacrificing one persistence dimension for improved performance. In FlashCubical, this is done by skipping the middle dimension, $H_1$. CubicalRipser can skip the computation of $H_2$. The output of the algorithms is not the same: this is to be considered when comparing their performance gains relative to computing the full 3 dimensional persistence. The results are given in Table~\ref{tab:reduced_3d}. The performance gains relative to computing the three dimensions are significative for FlashCubical on synthetic data, but are less so in the other cases.

\begin{table}[H]
\centering
\small
\setlength{\tabcolsep}{6pt}
\caption{3D reduced-mode runtime comparison}
\label{tab:reduced_3d}
\begin{tabular}{lrr}
\toprule
Case & flash\_no\_h1 Time (s) & cripser\_h01 Time (s) \\
\midrule
Synthetic 3D large ($128^3$) & 0.7 & 7.3 \\
Bonsai ($128^3$)             & 0.4 & 2.9 \\
Aneurism ($128^3$)           & 0.3 & 2.6 \\
\bottomrule
\end{tabular}
\end{table}

The Gudhi framework has a dedicated module, CubicalPersistence, that can notably compute 2D T-filtrations very efficiently (but not V-filtrations or 3D data). A user interested in getting the persistence of an image quickly, who is indifferent to choosing a T or V-filtration, might be interested in comparing it to FlashCubical. The results are in Table~\ref{tab:gudhi_t_reference}.

\begin{table}[H]
\centering
\small
\setlength{\tabcolsep}{6pt}
\caption{2D comparaison against GUDHI's fast cubical persistence module}
\label{tab:gudhi_t_reference}
\resizebox{\textwidth}{!}{%
\begin{tabular}{lrrrr}
\toprule
Case & Flash Time (s) & GUDHI-T Time (s) & Flash MB & GUDHI-T MB \\
\midrule
Synthetic 2D medium ($316 \times 316$)        & 0.02 & \textbf{0.01} & \textbf{3}  & 3  \\
Lena ($256 \times 256$)                       & 0.01 & \textbf{0.01} & \textbf{1}  & 1.5  \\
Synthetic 2D large ($1024 \times 1024$)       & 0.11 & \textbf{0.06} & \textbf{23} & 26 \\
DIV2K ($1024 \times 1024$, avg.\ $n=5$)       & 0.09 & \textbf{0.05} & 22 & \textbf{21} \\
\bottomrule
\end{tabular}%
}
\end{table}

GUDHI's cubical persistence is twice as fast for the same memory cost. As such, if both T or V-filtrations are viable and a user wishes to compute the persistance of a 2D image, GUDHI's cubical persistence is the most efficent. 

\section{Conclusion and future work}
\label{sec:conclusion}

We have presented Flash Cubical, a highly efficient implementation of persistent homology for V-filtrations filtrations on regular 2D and 3D cubical complexes induced by images over $\mathbb{F}_2$, built around three core ideas: using duality to compute top-dimensional persistence via union-find, locally pruning zero-persistence and already-classified cells before union-find computation, and pre-computing all local information in a lookup table to take advantage of the regularity of cubical complexes.

The benchmarks show consistent and significant improvements over other efficient implementations, both in terms of time and memory costs, with FlashCubical's strongest advantage being in 3D, particularly on real volumes.

\paragraph{T-filtration.}
The most natural extension of this work is to T-filtrations
. All three ideas extend naturally: the duality argument applies analogously, the pruning strategy is unchanged, and the lookup table extends via a pseudo-lower-star construction around each cubical cell.

\paragraph{Alpha filtrations.}
The duality and pruning tricks may extend to simplicial complexes under alpha filtrations. In that setting, apparent pairs \citep{bauer2021, zhang2020} provide a local zero-persistence detection mechanism analogous to the pruning strategy described here. Whether the full combination of tricks translates into practical gains for alpha filtrations remains an open question.

\paragraph{Further directions.}
Additional open directions include: extension to higher-dimensional cubical complexes, GPU parallelisation (potentially of the lookup and union-find phases), multi-threading to compute top and bottom dimension in parallel, a more efficient 3D $H_1$ algorithm, and further theoretical work on the elements discussed in this article. 

\bibliographystyle{plainnat}
\bibliography{references}

\appendix
\section{Elements of Proof}
\label{app:proof}
\subsection{Proposition and Setting}

\begin{quote}
If cohomology applies and top dimensional facets  have
one or two co-facets, then over $\mathbb{Z}/p\mathbb{Z}$ with $p$ prime, persistence can be computed by union find over a dual graph.
\end{quote}

In a $d+1$ dimensional complex, let 
\[
e_1,\ldots,e_n
\]
be the $d$-dimensional facets (so edges in 2D, squares in 3D), and let
\[
V_1,\ldots,V_k
\]
be the $d+1$-dimensional co-facets (so squares in 2D, cubes in 3D).

We place ourselves in the cohomology/dual setting, so the filtration ordering is reversed compared to the homology setting. The indexing used is the ordering in the cohomology setting: $e_1$ is first in the cohomology filtration, so last in the homology filtration. A facet $e_i$ being negative thus means being negative in the cohomology or dual sense, i.e. being positive in homology. 

To compute persistence on the top dimension, for any \(i \in \{1,\ldots,n\}\), we need to compute:

\begin{enumerate}
    \item Whether \(e_i\) is positive or not, i.e. whether
    \[
    \delta e_i \in \operatorname{Vect}(\delta e_j : j < i).
    \]
    with $\delta$ the co-boundary operator

    \item If \(e_i\) is negative, find \(V_l\), its positive pair.
\end{enumerate}

\subsection{First Observation}

\begin{observation}
Condition~(1) can be interpreted as a union-find connectivity computation over a dual graph.
\end{observation}

\paragraph{Sketch of argument.}

First, the format of $\delta e_i$:
the coboundary vector associated with a
top dimensional facet \(e_i\) has one of two forms.

If \(e_i\) has two incident top-dimensional co-facets \(V_r\) and \(V_s\), then
\[
\delta e_i = V_r - V_s.
\]
Equivalently,
\[
\delta e_i =
(0,\ldots,0,\underbrace{1}_{r},0,\ldots,0,
\underbrace{p-1}_{s},0,\ldots,0).
\]
All other coordinates are zero.

If \(e_i\) has only one incident top-dimensonial co-facet \(V_r\), then
\[
\delta e_i = V_r.
\]
Equivalently,
\[
\delta e_i =
(0,\ldots,0,\underbrace{1}_{r},0,\ldots,0).
\]
Again, all other coordinates are zero.

We have
\[
\delta e_i \in (\mathbb{Z}/p\mathbb{Z})^k.
\]
We define
\[
\phi : (\mathbb{Z}/p\mathbb{Z})^k
\longrightarrow
(\mathbb{Z}/p\mathbb{Z})^{k+1}
\]
by
\[
(x_1,\ldots,x_k)
\longmapsto
(x_1, \ldots,x_k, -x_1 - x_2 - \cdots - x_k).
\]

The map \(\phi\) is injective and linear. Therefore,
\[
\delta e_i \in \operatorname{Vect}(\delta e_j : j < i)
\]
if and only if
\[
\phi(\delta e_i)
\in
\operatorname{Vect}(\phi(\delta e_j) : j < i).
\]

Consequently, we can add an extra dimension to \(\delta e_i\), set to \(0\) if \(\delta e_i\) has two components, and set to \(p-1\) if \(\delta e_i\) only has one component, and condition~(1) stays equivalent. Adding this extra dimension can be seen as adding a "special" co-facet \(v_{k+1}\) that appears in the co-boundary vector of many facets.

\subsection{Union-Find Interpretation}

After this embedding, every relevant vector has the simple form
\[
C_i = V_r - V_s
\quad \in (\mathbb{Z}/p\mathbb{Z})^{k+1},
\]
for some indices \(r\) and \(s\). Equivalently, \(C_i\) has exactly two nonzero coordinates: a \(1\) in position \(r\), a \(p-1\) in position \(s\), and zeros elsewhere.

Thus the original linear dependence test
\[
\delta e_i \in \operatorname{Vect}(\delta e_j : j < i)
\]
is transformed into the test
\[
C_i \in \operatorname{Vect}(C_j : j < i),
\]
where all vectors \(C_j\) have this two-sparse form.

Each such vector \(C_j = V_r - V_s\) can be interpreted as an edge between the vertices \(r\) and \(s\) of a dual graph. Therefore,
\[
C_i \in \operatorname{Vect}(C_j : j < i)
\]
holds precisely when the two endpoints of the edge associated with \(C_i\) are already connected by previously inserted edges.

This leads to the "dual graph", where $d+1$ dimensional cofacets map to vertices and $d$ dimensional facets map to edges. Solving this span problem is then equivalent to doing a classical union-find, where edges are introduced in the order of the cohomology filtration. Note that a special vertex is added to represent the extra co-facet that was introduced, and any edge with originally only one vertex is also connected to this special vertex.

\subsection{Pairing Rule in the Dual Graph}

\begin{observation}
The dual graph structure, with the convention that the special vertex comes first in the cohomology filtration, provides a way to identify the positive co-facet/dual vertex that is to be paired with a negative facet/dual edge.
\end{observation}

\paragraph{Sketch of argument.}
When a negative facet/dual edge is added, we pair it with its youngest cofacet/dual vertex.

When \(e_i\) is added, we now have
\[
\delta e_i = 0
\]
in the cohomology group. So, if \(V_r\), and \(V_s\) are now the two co-facets of \(e_i\), then
\[
V_r = -V_s.
\]

This means that the groups maintained by union-find are equivalent to homology groups, and for each homology group there is a single positive cofacet/dual vertex. 

So when we apply \(\operatorname{find}\) to the two extrema cofacets/dual vertex of a negative facet/dual edge \(e_i\), we are pairing \(e_i\) with the youngest of the two.

Note: It is important to have the special vertex be the youngest of cofacets, so it is never paired.

\end{document}